%% file: main.tex
\documentclass[10pt,twocolumn,letterpaper]{article}

\usepackage[pagenumbers]{cvpr} 

\input{preamble}

\definecolor{cvprblue}{rgb}{0.21,0.49,0.74}
\usepackage[pagebackref,breaklinks,colorlinks,allcolors=cvprblue]{hyperref}

\def\paperID{13241} 
\def\confName{CVPR}
\def\confYear{2026}

\title{Detecting AI-Generated Images via Distributional Deviations from Real Images}

\author{ 
Yakun Niu,
Yingjian Chen,
Lei Zhang\thanks{Corresponding author}
\\
Henan Key Laboratory of Big Data Analysis and Processing, \\School of Computer and Information Engineering, Henan University
\\
{\tt\small yingjianchen@henu.edu.cn, zhanglei@henu.edu.cn, ykniu@henu.edu.cn}
}

\begin{document}
\maketitle
\input{sec/0_abstract}    
\input{sec/1_intro}
\input{sec/2_relatedwork}
\input{sec/3_method}
\input{sec/4_experiments}
\input{sec/5_conclusion}
{
    \small
    \bibliographystyle{ieeenat_fullname}
    \bibliography{main}
}
\input{sec/6_appendix}


\end{document}

%% file: preamble.tex
\usepackage{algorithm}
\usepackage{algorithmic}
\usepackage{enumitem}
\usepackage{amsmath}
\usepackage{amssymb}
\usepackage{siunitx}
\usepackage{makecell}
\usepackage{multirow}
\usepackage{colortbl}
\usepackage[table]{xcolor}
\usepackage{booktabs}
\usepackage{pifont}
\usepackage{soul}

%% file: sec/0_abstract.tex
\begin{abstract}
The rapid advancement of generative models has significantly enhanced the quality of AI-generated images, raising concerns about misinformation and the erosion of public trust. Detecting AI-generated images has thus become a critical challenge, particularly in terms of generalizing to unseen generative models. Existing methods using frozen pre-trained CLIP models show promise in generalization but treat the image encoder as a basic feature extractor, failing to fully exploit its potential. In this paper, we perform an in-depth analysis of the frozen CLIP image encoder (CLIP-ViT), revealing that it effectively clusters real images in a high-level, abstract feature space. However, it does not truly possess the ability to distinguish between real and AI-generated images. Based on this analysis, we propose a \textbf{M}asking-based \textbf{P}re-trained model \textbf{F}ine-\textbf{T}uning (\textbf{MPFT}) strategy, which introduces a Texture-Aware Masking (TAM) mechanism to mask textured areas containing generative model-specific patterns during fine-tuning. This approach compels CLIP-ViT to attend to the “distributional deviations” from authentic images for AI-generated image detection, thereby achieving enhanced generalization performance. Extensive experiments on the GenImage and UniversalFakeDetect datasets demonstrate that our method, fine-tuned with only a minimal number of images, significantly outperforms existing approaches, achieving up to 98.2\% and 94.6\% average accuracy on the two datasets, respectively.
\end{abstract}

%% file: sec/1_intro.tex
\section{Introduction}
\label{sec:intro}

\begin{figure}[t]
\centering
\includegraphics[width=1.0\linewidth]{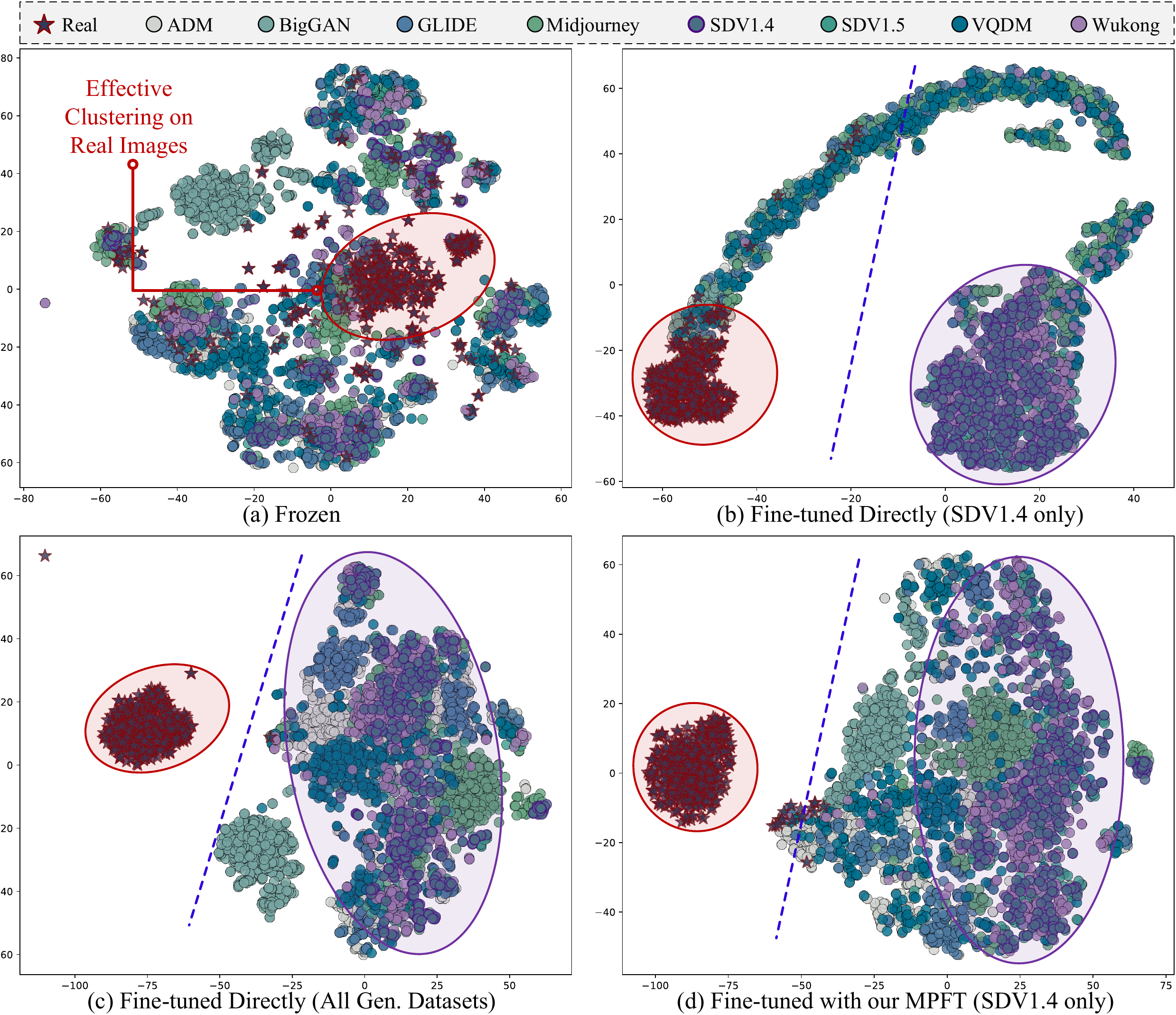}
\caption{t-SNE~\cite{van2008visualizing} visualization of features extracted by pre-trained CLIP-ViT from images generated by eight models in the GenImage dataset.
    (a) Frozen CLIP-ViT.
    (b) CLIP-ViT fine-tuned with only SDV1.4-generated images.
    (c) CLIP-ViT fine-tuned with images generated by \textbf{all involved generative models}.
    (d) CLIP-ViT fine-tuned with our proposed MPFT using \textbf{only SDV1.4-generated images}.}
\label{overview}
\end{figure}

In recent years, the rapid advancement of generative models~\cite{ho2020denoising, rombach2022high} has significantly improved image generation quality, making it easier to create highly realistic and high-quality images. For example, online platforms such as MidJourney \cite{midjourney} allow users to generate convincingly realistic and deceptive images easily. The potential misuse of these models raises serious concerns about the spread of misinformation. Consequently, detecting AI-generated images has become a critical challenge in preserving trust in digital content.

Many existing methods~\cite{luo2021generalizing, jeong2022bihpf, jeong2022frepgan, ju2023glff, deng2023new} achieve promising results in detecting images from seen generative models. However, they often struggle to generalize effectively to unseen generative models. To address this limitation, frozen pre-trained CLIP image encoder (CLIP-ViT)~\cite{radford2021learning} has been widely used for AI-generated image detection~\cite{ojha2023towards, koutlis2024leveraging, khan2024clipping, liu2024forgery, tan2025c2p}, demonstrating improved generalization. Nevertheless, these approaches typically treat the frozen CLIP-ViT as a simple feature extractor, lacking in-depth analysis and failing to fully exploit its potential. In this work, we conduct a visualization-based analysis of the pre-trained CLIP-ViT in the context of AI-generated image detection, uncovering key insights into its feature space that remain unexplored.

\subsection*{Feature Space Analysis of the Frozen Pre-trained CLIP Image Encoder} 
For an in-depth exploration of the features extracted by the frozen pre-trained CLIP-ViT~\cite{radford2021learning} for classification, we perform a feature space visualization using the GenImage~\cite{zhu2024genimage} dataset. The results show that the frozen CLIP-ViT effectively clusters real images (Figure~\ref{overview}, a) in a high-level feature space, benefiting from its strong representation capacity for real images acquired during pre-training on a large-scale dataset of 400 million real images. In this context, most AI-generated images appear as outliers, deviating from the feature distribution of real images. This enables the model to achieve generalized detection across diverse generative models by capturing a universal ``distribution shift''—that is, the deviation of generated image features from the compact cluster formed by real images in the high-dimensional feature space—rather than relying on specific low-level forgery artifacts~\cite{tan2024frequency, yan2024sanity, liu2024forgery, Chen_2025_ICCV}.
However, since the pre-trained encoder is not specifically optimized for this task, as reflected by the scattered distribution of AI-generated images, it lacks the ability to directly identify generated content and instead relies solely on the distribution gap for discrimination, which limits its detection accuracy in the absence of task-specific fine-tuning. This naturally raises a straightforward question: \ul{Why not fine-tune the CLIP-ViT directly to better adapt to the AI-generated image detection task?}

\subsection*{Limitations of Fine-tuning the Pre-trained CLIP Image Encoder for AI-Generated Image Detection} 
To answer this, we analyze the impact of fine-tuning on the pre-trained CLIP image encoder using the same GenImage dataset as in the previous analysis. Specifically, we visualize the feature space of CLIP-ViT fine-tuned on a single generative model (SDV1.4) and on a combination of eight generative models. After being fine-tuned on SDV1.4, the model shows clear clustering and separation between real images and SDV1.4-generated images (Figure~\ref{overview}, b). However, it exhibits overfitting, unable to distinguish images generated by other unseen models. This overfitting arises because the fine-tuning process encourages the model to rely on the differences between SDV1.4-generated images and real images for classification, while such unique differences~\cite{corvi2023intriguing, ojha2023towards} do not exist in images generated by other models, thus limiting its generalization ability. In contrast, when fine-tuned on images generated by all eight models, CLIP-ViT effectively clusters real images and distinguishes them from those generated by all seen models.

\begin{figure}[t]
\centering
    \includegraphics[width=1.0\linewidth]{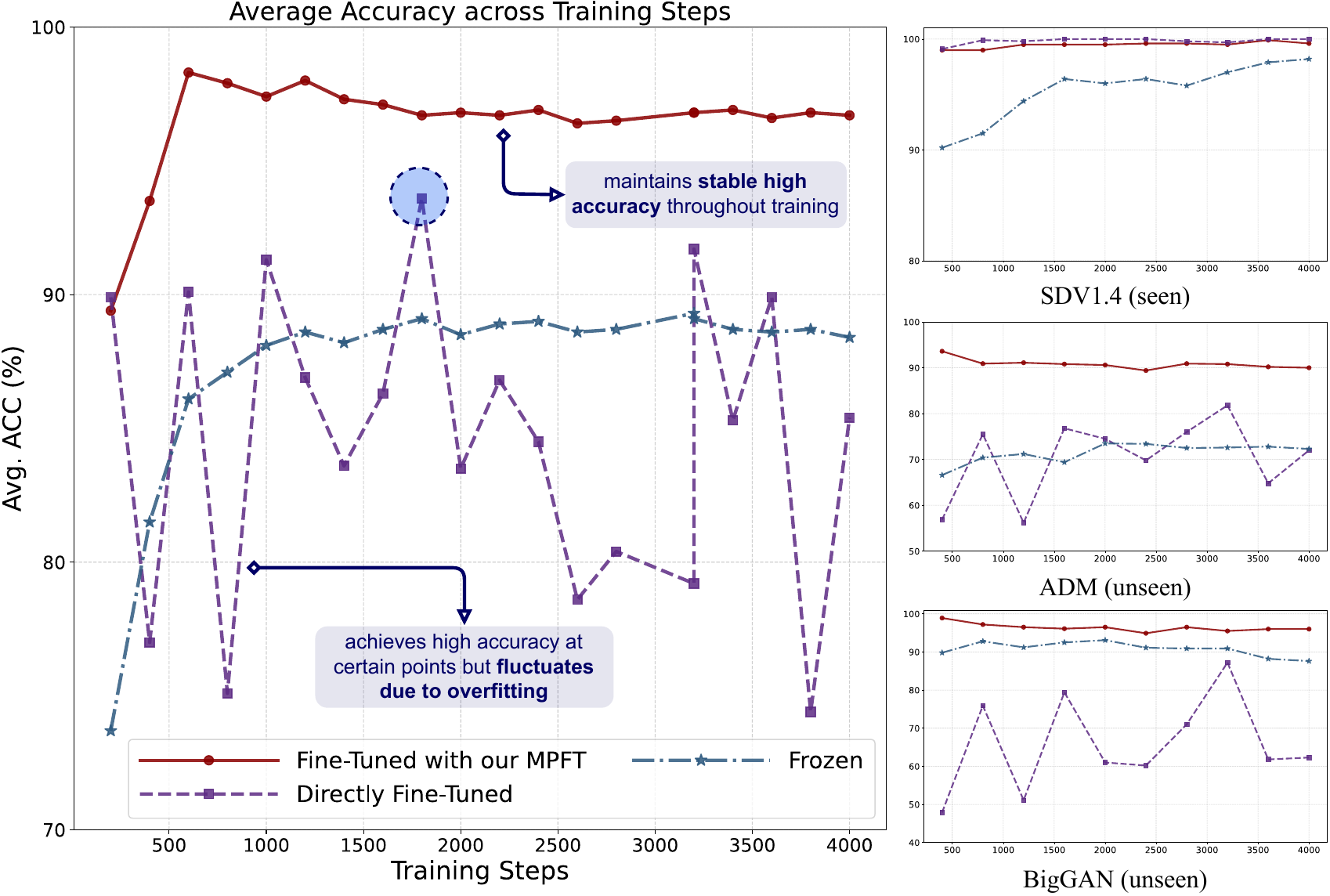}
    \caption{Left: Generalization performance (average accuracy across all 8 subsets of the GenImage test set) of three models: frozen CLIP-ViT with a trainable linear classifier, directly fine-tuned CLIP-ViT, and CLIP-ViT fine-tuned using our MPFT strategy. All models are trained on the SDV1.4 subset of the GenImage training set. Right: Accuracy of the models on the SDV1.4 (seen), ADM (unseen), and BigGAN (unseen) subsets of the GenImage test set.}
    \label{fig:line_chart}
\end{figure}

Additionally, to further investigate the fine-tuning process, we analyze the generalization performance of the model during fine-tuning on SDV1.4, as shown in Figure~\ref{fig:line_chart}. The results demonstrate that while the directly fine-tuned pre-trained CLIP-ViT achieves relatively good performance at certain stages, its generalization ability fluctuates and gradually degrades as fine-tuning progresses (purple line). In contrast, the frozen CLIP-ViT with a trainable linear classifier~\cite{ojha2023towards} maintains stable performance throughout (blue line). Similarly, the directly fine-tuned model shows steadily increasing accuracy on the SDV1.4 (seen), but exhibits fluctuating and limited performance on the ADM and BigGAN (both unseen). These observations confirm that: (1) The feature space of the pre-trained CLIP-ViT is compromised due to overfitting to the training data as fine-tuning progresses, causing fluctuations and performance degradation; and (2) The optimal generalization performance is typically reached in the early stage of fine-tuning (within 1,000 steps), suggesting that fine-tuning with the pre-trained CLIP-ViT requires only a small amount of data to achieve peak performance.

\subsection*{Motivation} 
Based on the above analysis, our motivation is to suppress the model’s reliance on specific forgery artifacts for detection during fine-tuning, while preserving its ability to identify real images based on the ``distributional deviations'' in the high-dimensional feature space. To achieve this, we propose masking texture-rich regions, which are areas that generative models typically struggle to synthesize realistically~\cite{zhong2023patchcraft} and that often contain model-specific patterns. For real images, CLIP-ViT learns highly abstract representations rather than relying on specific local textures, so the remaining regions retain sufficient information to effectively cluster real-image features. For AI-generated images, masking model-specific patterns prevents overfitting and encourages the model to rely on the inherent ``distributional deviations" between generated and real images across different models. This shifts the fine-tuning objective from capturing the differences between real and generated images in the training set to detecting the deviations from real images in the high-level feature space, thereby exploiting CLIP’s inherent ability to cluster real images and enabling generalized detection across diverse generative models.

Building on this motivation, we propose a simple yet effective \textbf{M}asking-based \textbf{P}re-trained model \textbf{F}ine-\textbf{T}uning strategy (\textbf{MPFT}) in this work. Specifically, we fine-tune the pre-trained CLIP-ViT with a small amount of data and introduce a Texture-Aware Masking (TAM) mechanism that selectively masks out local regions with rich textures during fine-tuning. As shown in Figure~\ref{overview} (c) and (d), our method demonstrates strong generalization: the pre-trained CLIP-ViT fine-tuned on only SDV1.4-generated images using our MPFT strategy performs comparably to one fine-tuned directly on images from all involved generative models. Moreover, our method exhibits stable generalization throughout the fine-tuning process (Figure~\ref{fig:line_chart}, red line). Extensive experiments on both the GenImage~\cite{zhu2024genimage} and UniversalFakeDetect~\cite{ojha2023towards} datasets demonstrate the effectiveness of our approach.

Our main contributions are three-fold as follows:
\begin{itemize}[noitemsep, topsep=1pt]
    \itemsep 0em
    \item \textbf{In-depth Analysis of the Frozen Pre-Trained CLIP-ViT.} 
    We perform an analysis and provide the insight that the frozen pre-trained CLIP-ViT achieves strong generalization through effective real image clustering, while also highlighting the necessity and limitations of fine-tuning.
    \item \textbf{A Novel Perspective on AI-generated Image Detection.}  
    Unlike previous methods that rely on low-level artifacts, our method emphasizes ``distributional deviations'' in high-dimensional feature space, achieving superior generalization across diverse generative models.
    \item \textbf{Effective Fine-tuning strategy.}  
    We propose MPFT, a fine-tuning strategy incorporating a Texture-Aware Masking (TAM) mechanism that masks rich-textured regions, mitigates overfitting, and preserves CLIP-ViT’s intrinsic ability to cluster real images effectively.
\end{itemize}

%% file: sec/2_relatedwork.tex
\section{Related Works}

\subsection{Feature Learning-Based Methods}
Previous methods based on spatial \cite{liu2020global, wang2020cnn, ju2023glff, deng2023new} and frequency \cite{frank2020leveraging, luo2021generalizing, jeong2022bihpf, jeong2022frepgan, tan2024frequency} domains demonstrated high accuracy in detecting images from the seen generative models but struggled to identify images generated by unseen models \cite{cozzolino2018forensictransfer, zhang2019detecting}. Recent methods have shifted towards identifying generalizable features that span various generative models. Specifically, LNP~\cite{liu2022detecting} discovered that noise patterns in real images exhibit consistent characteristics in the frequency domain, unlike generated images. LGrad~\cite{tan2023learning} transforms images into gradient representations. PatchCraft~\cite{zhong2023patchcraft} captures the residual pixel fluctuations between textured and smooth areas. NPR~\cite{tan2024rethinking} identifies artifacts from the upsampling process. DIRE~\cite{wang2023dire} and $\text{LaRE}^{2}$~\cite{luo2024lare} utilize reconstruction errors from diffusion models. In contrast to methods that rely on low-level artifacts, we introduce a novel perspective by focusing on the distributional deviations between real and AI-generated images in high-level, abstract feature representations, thereby achieving superior generalization.

\subsection{Pre-trained CLIP-based Methods}
Another line of research leverages large pre-trained models such as CLIP to enhance generalization in AI-generated image detection. For instance, Ojha~\cite{ojha2023towards} proposes using a pre-trained CLIP image encoder with a trainable linear classifier, which helps prevent overfitting and improves generalization. RINE~\cite{koutlis2024leveraging} extends this approach by utilizing intermediate-layer features from the frozen CLIP-ViT encoder. FatFormer~\cite{liu2024forgery} introduces an adapter to identify local forgery regions and incorporates a contrastive objective between image features and textual prompts. C2P-CLIP~\cite{tan2025c2p} injects category-related concepts into the image encoder by integrating the category common prompt. However, these methods generally treat the CLIP image encoder as a feature extractor. In contrast, we conduct a deeper investigation into the intrinsic properties of the pre-trained visual encoder and find that CLIP-ViT naturally clusters real images effectively in a high-level feature space. Based on this observation, our method explicitly exploits this property by fine-tuning the model to identify distributional deviations from real images. This novel perspective, which detects fake images by focusing on their distributional deviations from real images, significantly enhances generalization across diverse generative models.

%% file: sec/3_method.tex
\begin{figure*}[t]
	\centering
	\includegraphics[width=1.0\linewidth]{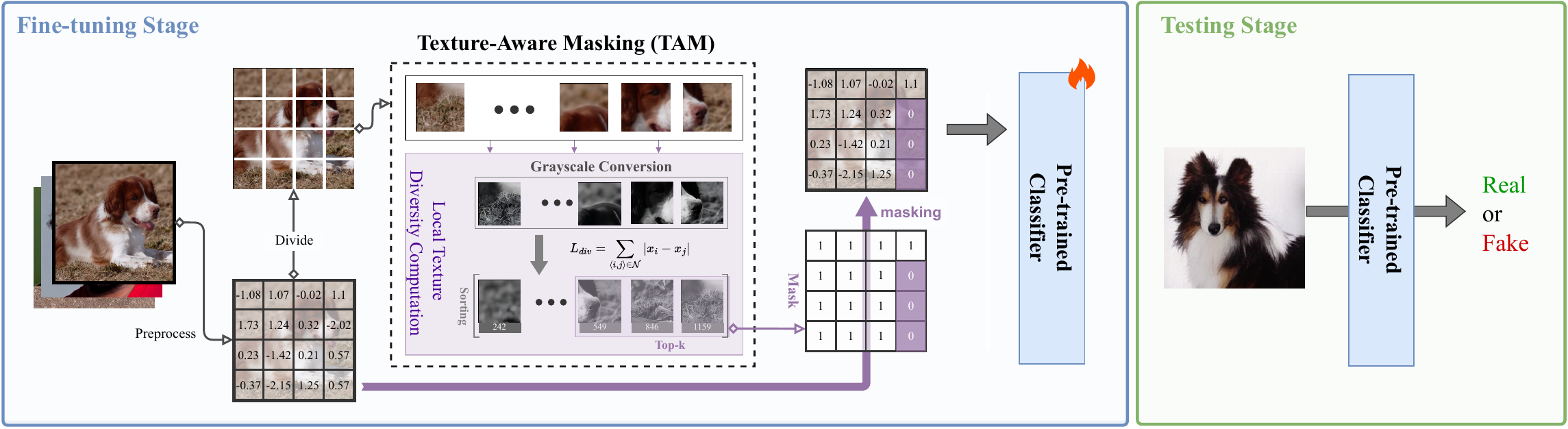}
	\caption{Overview of the Fine-tuning and Testing Pipeline.}
	\label{pipeline}
\end{figure*}

\section{Method}
In this section, we introduce \textbf{M}asking-based \textbf{P}re-trained model \textbf{F}ine-\textbf{T}uning strategy (\textbf{MPFT}), a simple yet effective masking-based pre-trained model fine-tuning strategy. MPFT employs a Texture-Aware Mask (TAM) to mask texture-rich regions—areas where generative models often struggle to generate intricate details, leaving behind model-specific patterns. By masking these regions, MPFT prevents the model from overfitting to such patterns during fine-tuning, encouraging it to focus on the ``distributional deviations'' between real and generated images. This enhances generalization and effectively leverages the pre-trained model’s inherent ability to cluster real images.

\subsection{Texture-Aware Masking (TAM) Mechanism.}
\label{sec: mask_gen}
Given an RGB image \( X \in \mathbb{R}^{C \times H \times W} \), we first divide it into \( N \) non-overlapping patches of size \( p \times p \). For each patch, we convert it to grayscale and compute its local texture diversity \( L_{div} \)~\cite{zhong2023patchcraft}, defined as the sum of absolute differences between adjacent pixel values along four directions:
\begin{equation}
L_{div} = \sum_{\langle i, j \rangle \in \mathcal{N}} |x_i - x_j|
\end{equation}
where $\langle i, j \rangle$ denotes all adjacent pixel pairs within the patch $x$, and $\mathcal{N}$ includes neighbors in the horizontal, vertical, diagonal, and anti-diagonal directions. The full equation is provided in the Appendix~\ref{appendix:equation}.

Patches are then ranked according to their $L_{div}$ scores, and the top-$k$ patches with the highest texture complexity are masked by setting the corresponding regions in the binary mask $M \in \mathbb{R}^{1 \times H \times W}$ to zero, based on the masking ratio $r_{\text{mask}}$, while the remaining regions are kept as one. The generated mask $M$ is applied to the input image prior to model fine-tuning. The pseudo-code of the proposed masking mechanism process is provided in Algorithm~\ref{algorithm1}.

\subsection{Pipeline.}
The overview of the pipeline is shown in Figure~\ref{pipeline}. In the fine-tuning stage, the input image undergoes preprocessing, transforming it into a suitable tensor format for model input. Before feeding the image into the model, the TAM mechanism is applied to all three channels to suppress regions with high texture complexity, as shown in:
\begin{equation}\label{eqn-8} 
	\begin{aligned}
		&X_{tensor} = \text{Preprocess}(X) \\
		&X_{input} = \text{TAM}(X_{tensor}) \circ X_{tensor}
	\end{aligned}
\end{equation}
In this equation, $X_{tensor}$ represents the preprocessed image tensor, $\text{TAM}(X_{tensor})$ is the binary mask, and $X_{input}$ denotes the final input tensor fed into the model.
In the testing stage, after preprocessing, the image is directly fed into the fine-tuned model for prediction.


\begin{algorithm}[!b]
\caption{Texture-Aware Mask Generation}
\label{algorithm1}
\begin{algorithmic}[1]
\REQUIRE An RGB image $x$ of size $C \times H \times W$, masking ratio $r_{\text{mask}}$, patch size $p$
\ENSURE A mask $M$ of size $1 \times H \times W$
\STATE Initialize mask $M \leftarrow \mathbf{1}^{1 \times H \times W}$
\STATE Divide $x$ into $N$ non-overlapping patches of size $C \times p \times p$
\FOR{$k = 1$ to $N$}
    \STATE Convert the $k$-th patch $x_k$ to grayscale: $g_k \leftarrow \text{mean}(x_k, \text{channel})$
    \STATE Compute $L_{\text{div}}(k)$ as the sum of absolute differences between adjacent pixels in $g_k$ (horizontal, vertical, diagonal)
\ENDFOR
\STATE Sort patch indices in descending order of $L_{\text{div}}$: $I \leftarrow \text{argsort}(L_{\text{div}}, \text{descending})$
\STATE $n_{\text{mask}} \leftarrow \lfloor N \times r_{\text{mask}} \rfloor$
\STATE $S \leftarrow I[1 : n_{\text{mask}}]$ \hfill // Indices of selected patches to be masked
\FOR{each index $s$ in $S$}
    \STATE Compute $(i, j)$ as the patch’s top-left coordinate in $M$
    \STATE Set $M[:, i:i+p-1, j:j+p-1] \leftarrow 0$
\ENDFOR
\RETURN $M$
\end{algorithmic}
\end{algorithm}

\begin{figure*}[t]
\centering
\includegraphics[width=1.0\linewidth]{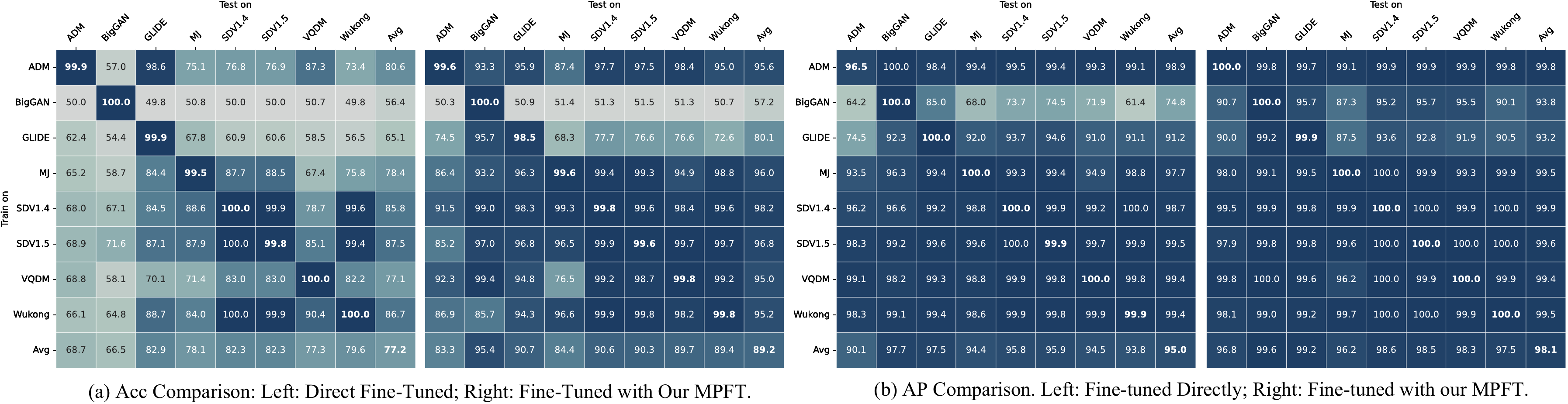}
\caption{Accuracy (Acc) Across 8 Subsets on GenImage: Comparison of CLIP-ViT with two fine-tuning strategies: direct fine-tuning (base) and fine-tuning with our proposed MPFT method. Models are fine-tuned on a single subset and evaluated on all 8 subsets. The color scale represents performance, with darker shades indicating higher accuracy.}
\label{img4}
\end{figure*} 

\begin{table*}[t]
\centering
\small
\setlength{\tabcolsep}{4.3pt}
\renewcommand{\arraystretch}{0.95}
\begin{tabular}{l l c c c c c c c c c}
    \toprule
    \multirow{2}{*}{Methods} & \multirow{2}{*}{Ref} & \multicolumn{8}{c}{Testing Subset} & \multirow{2}{*}{Avg. Acc.(\%)}\\
    \cmidrule(lr){3-10}
     & & ADM & BigGAN & GLIDE & MidJourney & SDV1.4 & SDV1.5 & VQDM & Wukong &  \\
    \midrule
    CNN-Spot & CVPR 2020 & 50.1 & 46.8 & 39.8 & 52.8 & 96.3 & 95.9 & 53.4 & 78.6 & 64.2 \\
    F3Net & ECCV 2020 & 49.9 & 49.9 & 50.0 & 50.1 & 99.9 & 99.9 & 49.9 & 99.9 & 68.7 \\
    Swin-T & ICCV 2021 & 49.8 & 57.6 & 67.6 & 62.1 & 99.9 & 99.8 & 62.3 & 99.1 & 74.8 \\
    CLIP-ViT & Base Model* & 68.0 & 67.1 & 84.5 & 88.6 & 100.0 & 99.9 & 78.7 & 99.6 & 85.8 \\
    $\text{LaRE}^2$ & CVPR 2024 & 61.7 & 68.7 & 88.5 & 74.0 & 100.0 & 99.9 & 97.2 & 100.0 & 86.3 \\
    UniFD & CVPR 2023 & 71.9 & 90.5 & 85.4 & 93.9 & 96.4 & 96.2 & 81.6 & 94.3 & 88.8 \\
    NPR & CVPR 2024 & 76.9 & 84.2 & 89.8 & 81.0 & 98.2 & 97.9 & 84.1 & 96.9 & 88.6 \\
    FreqNet & AAAI 2024 & 66.8 & 81.4 & 86.5 & 89.6 & 98.8 & 98.6 & 75.8 & 97.3 & 86.8 \\
    FatFormer & CVPR 2024 & 75.9 & 55.8 & 88.0 & 92.7 & 100.0 & 99.9 & 98.8 & 99.9 & 88.9 \\
    C2P-CLIP & AAAI 2025 & 96.4 & 98.7 & 99.0 & 88.2 & 90.9 & 97.9 & 96.5 & 98.8 & \underline{95.8} \\
    \midrule
    MPFT (CLIP-ViT)  & Ours & 91.5 & 99.0 & 98.3 & 99.3 & 99.8 & 99.6 & 98.4 & 99.6 & \textbf{98.2} \\
    \bottomrule
\end{tabular}
\caption{Comparison of Average Accuracy (Avg. Acc) between our method and existing methods on the GenImage test sets. Each model is trained on the SDV1.4 subset and evaluated across all test sets. The results for existing methods are directly cited from~\cite{tan2025c2p}. \textbf{Bold} and \underline{underline} represent the best and second-best performance, respectively.}
\label{tab:genimage}
\end{table*}

%% file: sec/4_experiments.tex
\section{Experimental Setup}
\subsection{Dataset}
To evaluate the effectiveness of our proposed method, we perform an evaluation on the two widely used benchmarks GenImage~\cite{zhu2024genimage} and UniversalFakeDetect~\cite{ojha2023towards}.

\subsubsection{GenImage.}
The GenImage dataset includes images generated by eight different generative models and spans a wide range of image resolutions. In our experiments, we follow the official dataset split by training on images generated by one model and evaluating generalization performance across all others. Specifically, we sample 800 training images from the original training set for fine-tuning.

\subsubsection{UniversalFakeDetect.}
The UniversalFakeDetect dataset contains images generated by 19 different generative models, including both GAN-based and diffusion-based models. Following prior work~\cite{liu2024forgery, tan2024frequency}, we train only on images generated by ProGAN and adopt a 4-class subset (car, cat, chair, horse) for training, selecting 800 images per class from the original training set.


\begin{table*}[ht]
    \centering
    \small
    \setlength{\tabcolsep}{2.15pt} 
    \renewcommand{\arraystretch}{1.05}  
    
    \begin{tabular}{l c c c c c c c c c c c c c c c c c c c c}
    \toprule
     \multirow{3}{*}{Method} & 
         \multicolumn{6}{c}{Generative Adversarial Networks} & 
         \multirow{3}{*}{\makecell{Deep\\fakes}} & 
         \multicolumn{2}{c}{Low level} & 
         \multicolumn{2}{c}{Perc. loss} & 
         \multirow{3}{*}{\makecell{Gui\\ded}} & 
         \multicolumn{3}{c}{LDM} & 
         \multicolumn{3}{c}{Glide} & 
         \multirow{3}{*}{Dalle} & 
         \multirow{3}{*}{\makecell{Avg.\\Acc}} \\
         
         \cmidrule(lr){2-7}
         \cmidrule(lr){9-10}
         \cmidrule(lr){11-12}
         \cmidrule(lr){14-16}
         \cmidrule(lr){17-19}

         & \makecell{Pro\\GAN} & \makecell{Cycle\\GAN} & \makecell{Big\\GAN} & \makecell{Style\\GAN} & \makecell{Gau\\GAN} & \makecell{Star\\GAN} & 
         & SITD & SAN & CRN & IMLE & 
         & \makecell{200\\steps} & \makecell{200\\w/cfg} & \makecell{100\\steps} & 
         \makecell{100\\27} & \makecell{50\\27} & \makecell{100\\10} & 
         & \\
        \midrule
        CNN-Spot & 100.0 & 85.2 & 70.2 & 85.7 & 79.0 & 91.7 & 53.5 & 66.7 & 48.7 & 86.3 & 86.3 & 60.1 & 54.0 & 55.0 & 54.1 & 60.8 & 63.8 & 65.7 & 55.6 & 69.6 \\
        Patchfor & 75.0 & 69.0 & 68.5 & 79.2 & 64.2 & 63.9 & 75.5 & 75.1 & 75.3 & 72.3 & 55.3 & 67.4 & 76.5 & 76.1 & 75.8 & 74.8 & 73.3 & 68.5 & 67.9 & 71.2 \\
        F3Net & 99.4 & 76.4 & 65.3 & 92.6 & 58.1 & 100.0 & 63.5 & 54.2 & 47.3 & 51.5 & 51.5 & 96.2 & 68.2 & 75.4 & 68.8 & 81.7 & 83.3 & 83.1 & 66.3 & 71.3 \\
        $\text{CLIP-ViT}^{*}$ & 100.0 & 78.1 & 75.3 & 85.7 & 62.6 & 100.0 & 55.7 & 69.0 & 72.4 & 49.8 & 50.3 & 81.3 & 87.9 & 84.6 & 90.2 & 70.3 & 79.0 & 74.5 & 89.0 & 76.6 \\
        LGrad & 99.8 & 85.4 & 82.9 & 94.8 & 72.5 & 99.6 & 58.0 & 62.5 & 50.0 & 50.7 & 50.8 & 77.5 & 94.2 & 95.9 & 94.8 & 87.4 & 90.7 & 89.6 & 88.4 & 80.3 \\
        UniFD & 100.0 & 98.5 & 94.5 & 82.0 & 99.5 & 97.0 & 66.6 & 63.0 & 57.5 & 59.5 & 72.0 & 70.0 & 94.2 & 73.8 & 94.4 & 79.1 & 79.9 & 78.1 & 86.8 & 81.4 \\
        FreqNet & 97.9 & 95.8 & 90.5 & 97.5 & 90.2 & 93.4 & 97.4 & 88.9 & 59.0 & 71.9 & 67.4 & 86.7 & 84.6 & 99.6 & 65.6 & 85.7 & 97.4 & 88.2 & 59.1 & 85.1 \\
        NPR & 99.8 & 95.0 & 87.6 & 96.2 & 86.6 & 99.8 & 76.9 & 66.9 & 98.6 & 50.0 & 50.0 & 84.6 & 97.7 & 98.0 & 98.2 & 96.3 & 97.2 & 97.4 & 87.2 & 87.6 \\
        FatFormer & 99.9 & 99.3 & 99.5 & 97.2 & 99.4 & 99.8 & 93.2 & 81.1 & 68.0 & 69.5 & 69.5 & 76.0 & 98.6 & 94.9 & 98.7 & 94.4 & 94.7 & 94.2 & 98.8 & 90.9 \\
        RINE & 100.0 & 99.3 & 99.6 & 88.9 & 99.8 & 99.5 & 80.6 & 90.6 & 68.3 & 89.2 & 90.6 & 76.1 & 98.3 & 88.2 & 98.6 & 88.9 & 92.6 & 90.7 & 95.0 & 91.3\\
        C2P-CLIP & 100.0 & 97.3 & 99.1 & 96.4 & 99.2 & 99.6 & 93.8 & 95.6 & 64.4 & 93.3 & 93.3 & 69.1 & 99.3 & 97.3 & 99.3 & 95.3 & 95.3 & 96.1 & 98.6 & \underline{93.8} \\
        \midrule
        MPFT & 99.9 & 99.2 & 97.3 & 94.9 & 97.6 & 99.7 & 97.6 & 94.7 & 90.9 & 87.7 & 84.8 & 82.0 & 99.3 & 97.2 & 99.5 & 95.5 & 96.3 & 96.8 & 98.4 & \textbf{94.6} \\
    \bottomrule
    \end{tabular}
\caption{Accuracy (Acc) results of forgery detection methods on UniversalFakeDetect, covering both GANs and diffusion models. \textbf{Bold} and \underline{underline} represent the best and second-best performance, respectively.}
\label{tab:uniacc}
\end{table*}

\begin{table*}[ht]
    \centering
    \small
    \setlength{\tabcolsep}{1.95pt} 
    \renewcommand{\arraystretch}{1.05}  
    
    \begin{tabular}{l c c c c c c c c c c c c c c c c c c c c}
    \toprule
     \multirow{3}{*}{Method} & 
         \multicolumn{6}{c}{Generative Adversarial Networks} & 
         \multirow{3}{*}{\makecell{Deep\\fakes}} & 
         \multicolumn{2}{c}{Low level} & 
         \multicolumn{2}{c}{Perc. loss} & 
         \multirow{3}{*}{\makecell{Gui\\ded}} & 
         \multicolumn{3}{c}{LDM} & 
         \multicolumn{3}{c}{Glide} & 
         \multirow{3}{*}{Dalle} & 
         \multirow{3}{*}{\makecell{Avg.\\AP}} \\
         
         \cmidrule(lr){2-7}
         \cmidrule(lr){9-10}
         \cmidrule(lr){11-12}
         \cmidrule(lr){14-16}
         \cmidrule(lr){17-19}

         & \makecell{Pro\\GAN} & \makecell{Cycle\\GAN} & \makecell{Big\\GAN} & \makecell{Style\\GAN} & \makecell{Gau\\GAN} & \makecell{Star\\GAN} & 
         & SITD & SAN & CRN & IMLE & 
         & \makecell{200\\steps} & \makecell{200\\w/cfg} & \makecell{100\\steps} & 
         \makecell{100\\27} & \makecell{50\\27} & \makecell{100\\10} & 
         & \\
        \midrule
        CNN-Spot & 100.0 & 93.5 & 84.5 & 99.5 & 89.5 & 98.2 & 89.0 & 73.8 & 59.5 & 98.2 & 98.4 & 73.7 & 70.6 & 71.0 & 70.5 & 80.7 & 84.9 & 82.1 & 70.6 & 83.6 \\
        Patchfor & 80.9 & 72.8 & 71.7 & 85.8 & 66.0 & 69.3 & 76.6 & 76.2 & 76.3 & 74.5 & 68.5 & 75.0 & 87.1 & 86.7 & 86.4 & 85.4 & 83.7 & 78.4 & 75.6 & 77.7 \\
        F3Net & 100.0 & 84.3 & 69.9 & 99.7 & 56.7 & 100.0 & 78.8 & 52.9 & 46.7 & 63.4 & 64.4 & 70.5 & 73.8 & 81.7 & 74.6 & 89.8 & 91.0 & 90.9 & 71.8 & 76.9 \\
        LGrad & 100.0 & 94.0 & 90.7 & 99.9 & 79.4 & 100.0 & 67.9 & 59.4 & 51.4 & 63.5 & 69.6 & 87.1 & 99.0 & 99.2 & 99.2 & 93.2 & 95.1 & 94.9 & 97.2 & 86.4 \\
        UniFD & 100.0 & 98.1 & 94.5 & 86.7 & 99.3 & 99.5 & 91.7 & 78.5 & 67.5 & 83.1 & 91.1 & 79.2 & 95.8 & 79.8 & 95.9 & 93.9 & 95.1 & 94.6 & 88.5 & 90.1 \\
        $\text{CLIP-ViT}^{*}$ & 100.0 & 95.2 & 95.9 & 99.5 & 95.0 & 100.0 & 93.8 & 85.2 & 77.4 & 93.7 & 97.5 & 91.2 & 93.9 & 90.6 & 95.3 & 79.0 & 85.1 & 82.1 & 92.3 & 91.7 \\
        FreqNet & 99.9 & 99.6 & 96.1 & 99.9 & 99.7 & 98.6 & 99.9 & 94.4 & 74.6 & 80.1 & 75.7 & 96.3 & 96.1 & 100.0 & 62.3 & 99.8 & 99.8 & 96.4 & 77.8 & 92.0 \\
        NPR & 100.0 & 99.5 & 94.5 & 99.9 & 88.8 & 100.0 & 84.4 & 98.0 & 100.0 & 50.2 & 50.2 & 98.3 & 99.9 & 99.9 & 99.9 & 99.9 & 99.9 & 99.9 & 99.3 & 92.8 \\
        FatFormer & 100.0 & 100.0 & 100.0 & 99.8 & 100.0 & 100.0 & 98.0 & 97.9 & 81.2 & 99.8 & 99.9 & 92.0 & 99.8 & 99.1 & 99.9 & 99.1 & 99.4 & 99.2 & 99.8 & 98.2 \\
        RINE & 100.0 & 100.0 & 99.9 & 99.4 & 100.0 & 100.0 & 97.9 & 97.2 & 94.9 & 97.3 & 99.7 & 96.4 & 99.8 & 98.3 & 99.9 & 98.8 & 99.3 & 98.9 & 99.3 & \underline{98.8}\\
        C2P-CLIP & 100.0 & 100.0 & 100.0 & 99.5 & 100.0 & 100.0 & 98.6 & 98.9 & 84.6 & 99.9 & 100.0 & 94.1 & 100.0 & 99.8 & 100.0 & 99.7 & 99.8& 99.8 & 99.9 & 98.7 \\
        \midrule
        MPFT & 100.0 & 100.0 & 99.9 & 99.3 & 100.0 & 100.0 & 96.8 & 100.0 & 95.5 & 100.0 & 99.9 & 95.6 & 99.8 & 99.4 & 99.9 & 98.1 & 98.8 & 98.9 & 99.6 & \textbf{99.0} \\
    \bottomrule
\end{tabular}
\caption{Average Precision (AP) results of forgery detection methods on UniversalFakeDetect.}
\label{tab:uniap}
\end{table*}

\subsection{Baselines}
To comprehensively evaluate our method, we compare it against CLIP-ViT~\cite{radford2021learning} fine-tuned directly (base), as well as various existing forgery detection methods. These include CNN-Spot~\cite{wang2020cnn}, F3Net~\cite{qian2020thinking}, Patchfor~\cite{chai2020makes}, Swin-T~\cite{liu2021swin}, UniFD~\cite{ojha2023towards}, LGrad~\cite{tan2023learning}, NPR~\cite{tan2024rethinking}, $\text{LaRE}^2$, FreqNet~\cite{tan2024frequency}, RINE~\cite{koutlis2024leveraging}, FatFormer~\cite{liu2024forgery}, and C2P-CLIP~\cite{tan2025c2p}.

\subsection{Implementation Details}
For training, input images were randomly cropped to 224 × 224, with horizontal flipping and rotation applied for data augmentation. Images that did not meet the minimum crop requirements were expanded by stitching repeated content to achieve the necessary crop size. In contrast, only center cropping was applied during testing. The Adam optimizer \cite{kingma2014adam} with beta parameters (0.9, 0.999) was employed to minimize binary cross-entropy loss. All experiments were conducted on an Nvidia GeForce RTX 3090 GPU. More detailed Implementation details can be found in the Appendix~\ref{appendix: hyperparameter}.

\subsection{Evaluation metric}
In accordance with the protocols outlined in prior methods~\cite{ojha2023towards, luo2024lare, tan2025c2p}, we use Accuracy (Acc) and Average Precision (AP) as the primary evaluation metrics. To evaluate the model's generalization capability, we compute the average Acc (Avg. Acc) and average AP (Avg. AP) across datasets from different generative models.

\section{Results and Analysis}
\subsection{Evaluation of Performance over Base method}
To evaluate the effectiveness of our proposed method, we conducted comparative experiments on the GenImage dataset, comparing the base CLIP-ViT (fine-tuned directly) with the version fine-tuned using our method. The models were fine-tuned on one subset and evaluated across all eight subsets. Additionally, we evaluated the models fine-tuned on images generated by all eight generative models to assess their generalization capability further. As shown in Figure \ref{img4}, our method demonstrates exceptional generalization performance, achieving a 12.0\% improvement in average Acc and a 3.1\% improvement in average AP compared to the directly fine-tuned version. Among all subsets, the model fine-tuned on SDV1.4 achieves the best overall performance, while fine-tuning on BigGAN leads to limited generalization due to its low resolution (128x128) and clear separability from real images, in line with prior reports~\cite{zhu2024genimage, luo2024lare}.

\subsection{Comparison with Existing Methods}
To further demonstrate the effectiveness of our proposed method, we compared its performance with recent methods on both GenImage and UniversalFakeDetect datasets, following the previous method~\cite{tan2025c2p}.

\subsubsection{Evaluation on GenImage.}
We follow existing methods~\cite{zhu2024genimage, tan2025c2p} and use the SDV1.4 for fine-tuning. The results of accuracy (Acc) are shown in Table~\ref{tab:genimage}. Notably, our method achieves 98.2\% Avg. Acc using only 1\% of the training data, outperforming all existing methods and reaching a 2.4\% improvement in Avg.Acc compared to the current SOTA method, C2P-CLIP~\cite{tan2025c2p}. This demonstrates that our method achieves exceptional generalization by enabling the pre-trained CLIP-ViT to focus on the ``distributional deviations'' between real and AI-generated images in the high-dimensional feature space. 

\subsubsection{Evaluation on UniversalFakeDetect.}
Table~\ref{tab:uniacc} and Table~\ref{tab:uniap} present the accuracy (Acc) and average precision (AP) results. Our method outperforms all existing methods, achieving an Avg. Acc of 94.6\% and an Avg. AP of 99.0\%. Specifically, CLIP-ViT fine-tuned using our MPFT surpasses the directly fine-tuned version by 18.0\% in Avg. Acc and 7.3\% in Avg. AP. Moreover, our approach outperforms the latest SOTA method in both Avg. Acc and Avg. AP. These results further demonstrate the effectiveness of our approach.

\subsection{Ablation Study}
In this section, we present comprehensive ablation studies on both the GenImage~\cite{zhu2024genimage} and UniversalFakeDetect~\cite{ojha2023towards} datasets to analyze the impact of different fine-tuning data volumes and masking strategies. Additional experiments are in the Appendix~\ref{appendix:experiments}.

\begin{table}[t]
\centering
    \small
    \setlength{\tabcolsep}{3pt}
    \renewcommand{\arraystretch}{1.0}
    \begin{tabular}{l c c c c}
        \toprule
        \multirow{2}{*}{Masking Strategy} & \multicolumn{2}{c}{GenImage} & \multicolumn{2}{c}{UniversalFakeDetect} \\
        \cmidrule(lr){2-3} \cmidrule(lr){4-5}
         & Avg.Acc & Avg.AP & Avg.Acc & Avg.AP\\
        \midrule
        CutOut & 90.3 & 99.4 & 91.3 & 98.0 \\
        GridMask & 92.8 & 99.6 & 91.3 & 97.2 \\
        Random Masking & 91.1 & 99.6 & 93.1 & 98.1 \\
        \midrule
        TAM (Low) & 87.8 & 97.8 & 90.6 & 98.2 \\
        TAM (Both) & 93.2 & 99.6 & 92.5 & 98.4 \\
        \midrule
        TAM (Ours) & \textbf{98.2} & \textbf{99.9} & \textbf{94.6} & \textbf{99.0}\\
        \toprule
    \end{tabular}
\caption{Average accuracy and average AP evaluated on the GenImage and UniversalFakeDetect datasets under different masking strategies. Top-performing results are highlighted in bold. TAM (Low) masks regions with the least texture, whereas TAM (Both) masks both high- and low-texture regions equally, with the same total masking ratio.}
\label{tab:masking}
\end{table}

\begin{figure}[t]
\centering
\includegraphics[width=8.2cm]{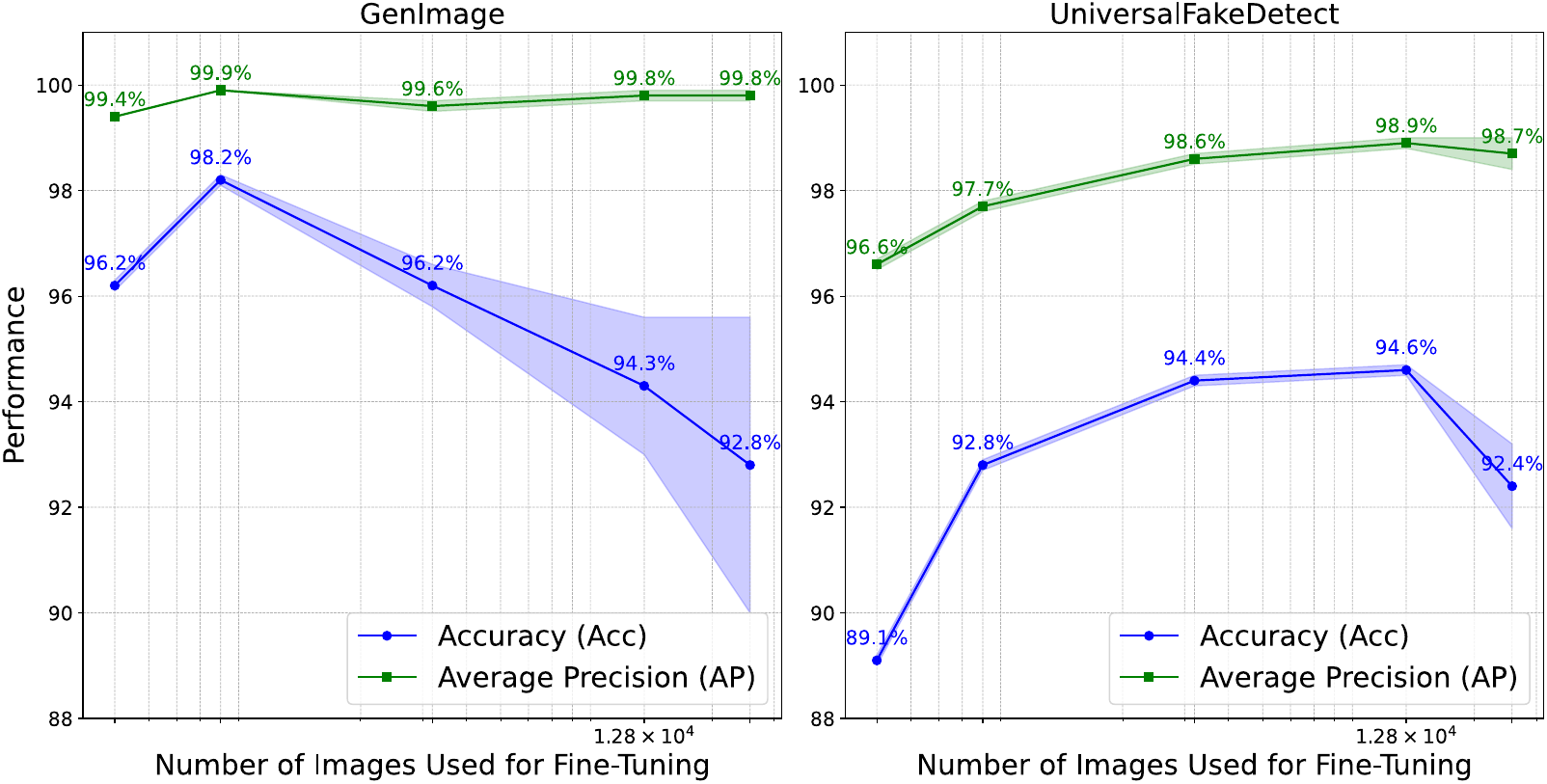}
\caption{Average Acc and average AP of CLIP-ViT fine-tuned with our method using different numbers of images (400, 800, 3200, 12800, and 25600)—with an equal split between real and generated images—evaluated on the GenImage and UniversalFakeDetect. To reflect the stability of the model’s generalization, we report the results averaged over the last five epochs.}
\label{line_count}
\end{figure}

\begin{figure*}[t]
	\centering
	\includegraphics[width=1.0\linewidth]{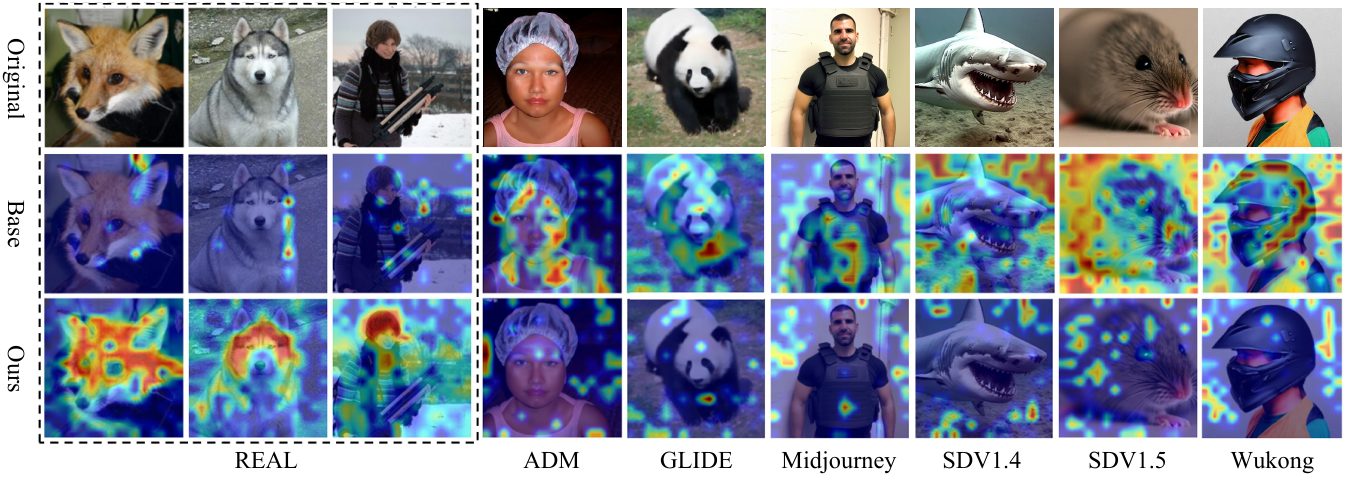}
	\caption{Class Activation Map (CAM) \cite{zhou2016learning} visualization for real and AI-generated images, comparing CLIP-ViT fine-tuned directly (base) with SDV1.4 and fine-tuned using our proposed MPFT.}
	\label{cam}
\end{figure*}


\subsubsection{Impact of Different Masking Strategies.}
To further evaluate our proposed fine-tuning framework, we investigate the effects of different masking strategies, as shown in Table~\ref{tab:masking}. We compare several widely used masking methods—CutOut~\cite{devries2017improved}, GridMask~\cite{chen2020gridmask}, and Random Erasing~\cite{zhong2020random}—with our proposed Texture-Aware Masking (TAM) under various masking configurations. The experimental results demonstrate that our fine-tuning strategy, MPFT, is compatible with all masking methods, each achieving satisfactory performance. However, our TAM consistently outperforms them. This suggests that while masking arbitrary regions of an image can constrain the model from overfitting generative model–specific patterns, explicitly masking texture-rich areas is more effective. Furthermore, the results of masking low-texture regions or jointly masking both high- and low-texture regions further confirm that models tend to rely on texture-rich areas to distinguish real images from AI-generated ones.

\subsubsection{Impact of Fine-tuning Data Volume.}
To investigate the impact of fine-tuning data volume, we analyze the performance of CLIP-ViT fine-tuned with varying data amounts using our method, as shown in Figure~\ref{line_count}. The results show that the pre-trained CLIP-ViT achieves optimal generalization with a small amount of fine-tuning data (800 samples for GenImage and 3200 or 12800 for UniversalFakeDetect). As the data volume increases, generalization performance becomes unstable and deteriorates. This is due to the model's tendency to overfit local forgery artifacts when exposed to more data during fine-tuning, shifting the decision criterion from "real image distribution" to "specific artifacts," which leads to overfitting and reduced generalization.

\subsection{Robustness Analysis}
To evaluate the robustness of our method under various perturbations, we applied noise, blurring, JPEG compression, and random cropping, with more details provided in Appendix~\ref{appendix:robustness}. The results, presented in Table~\ref{tab: robust}, show that noise and blurring significantly degrade model performance, with noise having the most pronounced effect. In contrast, random cropping and JPEG compression have minimal impact on performance, indicating that our method is more sensitive to image quality and exhibits strong robustness to geometric transformations. Additionally, to simulate real-world conditions, we sequentially applied all four perturbations—noise, blurring, JPEG compression, and random cropping—on both the training and test sets. Although our method’s performance degrades in this real-world scenario, it still maintains relatively high accuracy.

\begin{table}[t]
\centering
\newcommand{\cmark}{\ding{51}}
\newcommand{\xmark}{\ding{55}}
\small
\setlength{\tabcolsep}{1.3pt}
\renewcommand{\arraystretch}{1.0}
\begin{tabular}{c c c c l l}
    \toprule
    w/Blur & w/Cropping & w/JPEG & w/Noise & Avg.Acc.(\%) & Avg.AP.(\%) \\
    \midrule
    \multicolumn{6}{c}{GenImage} \\
    \midrule
    \cmark & \xmark & \xmark & \xmark & 81.1 (17.1$\downarrow$) & 87.7 (12.2$\downarrow$) \\
    \xmark & \cmark & \xmark & \xmark & 97.0 (1.2$\downarrow$) & 99.5 (0.4$\downarrow$)\\
    \xmark & \xmark & \cmark & \xmark & 89.2 (9.0$\downarrow$) & 3.2 (1.09$\downarrow$)\\
    \xmark & \xmark & \xmark & \cmark & 79.1 (19.1$\downarrow$)& 90.5 (9.4$\downarrow$)\\
    \cmark & \cmark & \cmark & \cmark & 84.9 (13.3$\downarrow$) & 89.3 (10.6$\downarrow$)\\
    \midrule
    \multicolumn{6}{c}{UniversalFakeDetect} \\
    \midrule
    \cmark & \xmark & \xmark & \xmark & 84.5 (10.1$\downarrow$) & 90.5 (8.5$\downarrow$) \\
    \xmark & \cmark & \xmark & \xmark & 92.1 (2.5$\downarrow$) & 98.4 (0.6$\downarrow$)\\
    \xmark & \xmark & \cmark & \xmark & 90.2 (4.4$\downarrow$) & 95.9 (3.1$\downarrow$)\\
    \xmark & \xmark & \xmark & \cmark & 81.3 (13.3$\downarrow$)& 91.4 (7.5$\downarrow$)\\
    \cmark & \cmark & \cmark & \cmark & 87.2 (7.4$\downarrow$) & 94.3 (4.7$\downarrow$)\\
    \bottomrule
\end{tabular}
\caption{Average Accuracy (Avg.Acc) and Average Precision (Avg.Ap) evaluated on UniversalFakeDetect under various perturbations.}
\label{tab: robust}
\end{table}

\subsection{Class Activation Map Visualization}
We conducted class activation map (CAM) analysis on both real and generated images to further investigate our method, as shown in Figure~\ref{cam}. The CLIP-ViT model fine-tuned directly on SDV1.4 responds more strongly to generated images, showing less attention to real images. This suggests that the directly fine-tuned model overfits to generative model-specific patterns within the generated images, focusing primarily on local artifacts, which diminishes its ability to generalize across different generative models. In contrast, when fine-tuning CLIP-ViT using our method, the model exhibits strong attention to real images and minimal response to AI-generated images. The visualization results further confirm that our method effectively guides the pre-trained CLIP-ViT to detect AI-generated images by referencing real images and recognizing deviations from their distributions in the highly abstract feature space, rather than relying on local forgery details.

%% file: sec/5_conclusion.tex
\section{Conclusion}
In this paper, we propose a simple yet effective Masking-based Pre-trained model Fine-Tuning (MPFT) strategy that enhances the generalization of AI-generated image detection. By introducing a Texture-Aware Masking (TAM) mechanism, we leverage the inherent strengths of the pre-trained CLIP-ViT to focus on the ``distributional deviations'' between real and generated images in a high-level feature space. Our method shifts the task from distinguishing real and generated images to identifying real images, which enables better generalization across a wide range of generative models. Extensive experiments on the GenImage and UniversalFakeDetect datasets demonstrate that our method outperforms existing approaches, showcasing exceptional generalization ability.

\noindent{\textbf{Limitations.}} Admittedly, our method is sensitive to the quality of the generated images used for training. If the training data contains lower-quality images, it can negatively impact its performance. In future work, we aim to develop more robust methods that can maintain strong performance even with lower-quality training images.

%% file: sec/6_appendix.tex
\clearpage
\appendix
\label{sec:appendix}

\section*{Appendix}

\section{Dataset Details}
\subsection{GenImage}
GenImage~\cite{zhu2024genimage} consists of subsets generated by 8 different generative models, which are split into training and testing sets. In this work, we have used only a small portion of the training data for fine-tuning. Detailed data can be found in Table~\ref{tab:genimage}.

\begin{table}[h]
    \centering
    \small
    \setlength{\tabcolsep}{4.3pt}
    \renewcommand{\arraystretch}{0.9}
    \begin{tabular}{lccc}
        \toprule
        Generative Models & Train Count & Test Count & Neg\% \\
        \midrule
        ADM   & 800  & 3,200 & 50\% \\
        BigGAN & 800  & 3,200 & 50\% \\
        GLIDE  & 800 & 3,200 & 50\% \\
        MidJourney  & 800 & 3,200 & 50\% \\
        Stable Diffusion V1.4 & 800 & 3,200 & 50\% \\
        Stable Diffusion V1.5 & 800 & 3,200 & 50\% \\
        VQDM & 800 & 3,200 & 50\% \\
        Wukong & 800 & 3,200 & 50\% \\
        \bottomrule
    \end{tabular}
    \caption{Statistics of the GenImage Dataset. We report the training size and testing size of each subset, and the proportion of negative samples.}
    \label{tab:genimage}
\end{table}

\subsection{UniversalFakeDetect}
UniversalFakeDetect~\cite{ojha2023towards} includes image subsets generated by 19 different generative models, covering both GAN-based and diffusion-based methods. For our experiments, we use ProGAN as the training set and adopt a 4-class subset (car, cat, chair, horse) as in previous works~\cite{tan2023learning, liu2024forgery, tan2025c2p}. During fine-tuning, we randomly select a small portion of data from each class. Details of the dataset are provided in Table~\ref{tab:uni}.

\begin{table}[h]
    \centering
    \small
    \setlength{\tabcolsep}{3.5pt}
    \renewcommand{\arraystretch}{0.9}
    \begin{tabular}{lccc}
        \toprule
        Generative Models & Size & Class Count & Neg\% \\
        \midrule
        \textbf{Training Set} &&& \\
        \midrule
        ProGAN & 3,200 for each & car, cat, chair, horse & 50\% \\
        \midrule
        \textbf{Testing Set} &&& \\
        \midrule
        ProGAN   & 8,000  & 20 & 50\% \\
        CycleGAN & 2,642  & 6 & 50\% \\
        BigGAN  & 4,000 & N/A & 50\% \\
        StyleGAN  & 11,982 & 3 & 50\% \\
        GauGAN & 10,000 & N/A & 50\% \\
        StarGAN & 3,998 & N/A & 50\% \\
        Deepfakes & 5,405 & N/A & 49.9\% \\
        SITD & 360 & N/A & 50\% \\
        SAN & 438 & N/A & 50\% \\
        CRN & 12,764 & N/A & 50\% \\
        IMLE & 12,764 & N/A & 50\% \\
        Guided & 2,000 & N/A & 50\% \\
        LDM 200 steps & 3,000 & N/A & 33.3\% \\
        LDM 200 w/CFG & 3,000 & N/A & 33.3\% \\
        LDM 100 steps & 3,000 & N/A & 33.3\% \\
        Glide-100-27 & 3,000 & N/A & 33.3\% \\
        Glide-50-27 & 3,000 & N/A & 33.3\% \\
        Glide-100-10 & 3,000 & N/A & 33.3\% \\
        DALL-E & 3,000 & N/A & 33.3\% \\
        \bottomrule
    \end{tabular}
    \caption{Statistics of the UniversalFakeDetect Dataset. We report the size of each subset, the number of classes, and the proportion of negative samples.}
    \label{tab:uni}
\end{table}

\section{Full Equation for the Calculation of $L_{div}$}
\label{appendix:equation}

\begin{align*}
L_{div} = & \sum_{i=1}^{M} \sum_{j=1}^{M-1} |x_{i,j} - x_{i,j+1}| \\
              & + \sum_{i=1}^{M-1} \sum_{j=1}^{M} |x_{i,j} - x_{i+1,j}| \\
              & + \sum_{i=1}^{M-1} \sum_{j=1}^{M-1} |x_{i,j} - x_{i+1,j+1}| \\
              & + \sum_{i=1}^{M-1} \sum_{j=1}^{M-1} |x_{i+1,j} - x_{i,j+1}|.
\end{align*}



\section{Additional Experiments}
\label{appendix:experiments}
In this section, we present additional experiments, including the impact of masking parameters and the influence of different training classes in ProGAN on generalization performance.

\subsection{Effect of Masking Ratio $r_{\text{mask}}$ and Patch Size $p$}
We conducted ablation experiments to evaluate the effects of different masking ratios $r_{\text{mask}}$. To increase data variability during training, the masking ratio was randomly selected within a specified range. The results are presented in Table~\ref{table4}. On GenImage, a masking ratio covering 60\%-80\% of the image area yielded the best performance. For UniversalFakeDetect, a lower masking ratio achieved the best results. This is due to the relatively low image quality in the ProGAN training set of UniversalFakeDetect, as well as the smaller image size. Excessive masking impairs feature extraction, leading to performance degradation.

\begin{table}[h]
\centering
\small
    \setlength{\tabcolsep}{3.5pt}
    \renewcommand{\arraystretch}{1.0}
    \begin{tabular}{c|c c c c}
        \toprule
        \multirow{3}{*}{\makecell{Masking Ratio \\ $r_{mask}$}} & \multicolumn{2}{c}{GenImage} & \multicolumn{2}{c}{UniversalFakeDetect} \\
        \cmidrule(lr){2-3} \cmidrule(lr){4-5}
        & Avg.Acc & Avg.AP & Avg.Acc & Avg.AP \\
        \midrule
        (0.0, 0.2) & 91.6 & 99.7 & 91.7 & 98.3 \\
        (0.0, 0.4) & 96.2 & 99.7 & \cellcolor{lightgray}94.6 & \cellcolor{lightgray}99.0 \\
        (0.0, 0.6) & 93.6 & 99.6 & 94.5 & 98.9 \\
        (0.0, 0.8) & 93.9 & 99.8 & 91.1 & 98.0 \\
        (0.0, 1.0) & 89.6 & 99.7 & 92.2 & 98.6 \\
        (0.2, 0.4) & 93.7 & 99.7 & 91.5 & 98.4 \\
        (0.4, 0.6) & 95.8 & 99.8 & 93.0 & 97.3 \\
        (0.6, 0.8) & \cellcolor{lightgray}98.2 & \cellcolor{lightgray}99.9 & 69.4 & 82.4 \\
        (0.8, 1.0) & 90.7 & 99.7 & 65.2 & 87.8 \\
        \bottomrule
\end{tabular}
\caption{Avg. Acc and Avg. AP on GenImage and UniversalFakeDetect, with variations in masking ratio.}
\label{table4}
\end{table}

\begin{figure}[h]
\centering
\includegraphics[width=8.2cm]{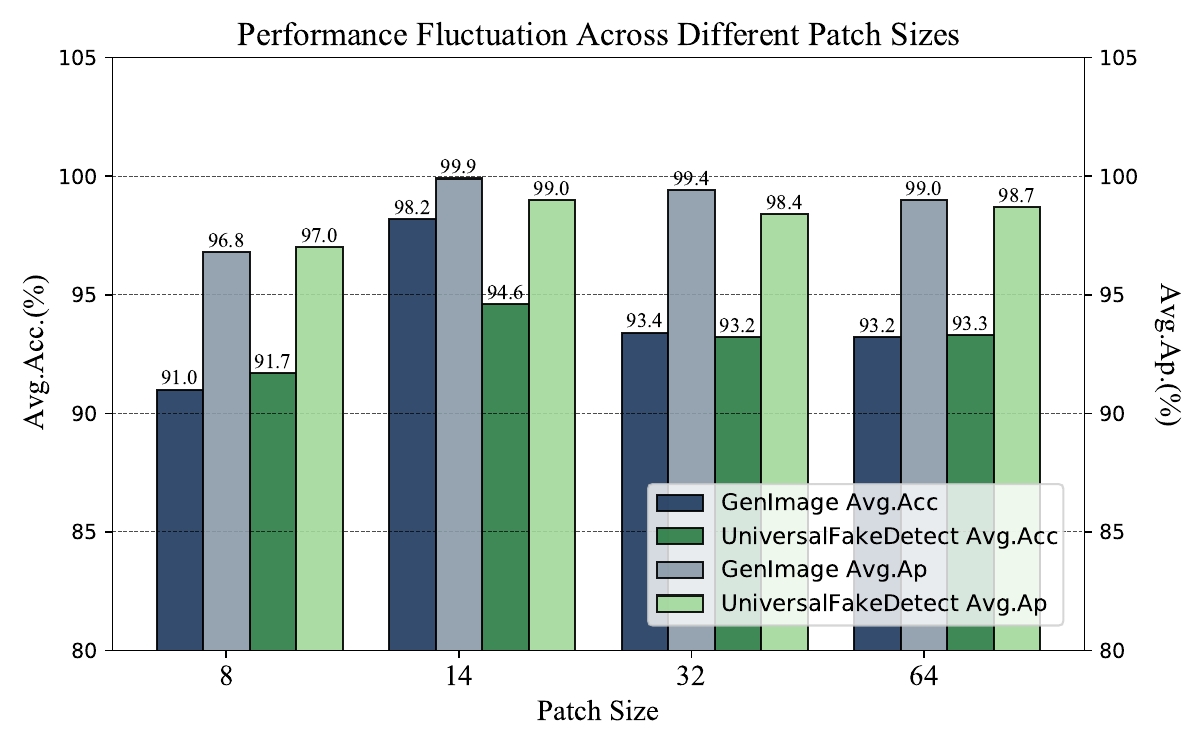}
\caption{Avg. Acc and Avg. AP on GenImage and UniversalFakeDetect, with variations in masking patch size.}
\label{barchart_patch}
\end{figure}

Additionally, for the masking patch size p, we initially set it to 14, matching the patch size used in the CLIP-ViT pre-training. However, we also conducted additional ablation studies to comprehensively investigate the influence of different patch sizes at the optimal masking ratio setting on GenImage and UniversalFakeDetect, as shown in Figure~\ref{barchart_patch}. The results demonstrate that a patch size of 14 achieves the best performance, as expected, while both larger and smaller patch sizes lead to performance degradation.

\section{Robustness Analysis}
\label{appendix:robustness}
In this section, we present robustness experiments to evaluate the effects of various perturbations on model performance, following the methodology outlined by Frank \textit{et~al.} \cite{frank2020leveraging}. These experiments are conducted on the GenImage \cite{zhu2024genimage} and UniversalFakeDetect.
\subsection{Perburbations}
\noindent{\textbf{Noise:}} Random Gaussian noise is added to the input images by selecting a variance value from a uniform distribution within the range [5.0, 20.0], controlling the noise intensity. This results in a noisy image with the same dimensions as the original, but with random variations in pixel intensity.

\noindent{\textbf{Blurring:}} Gaussian blur is applied to the input images using a randomly selected kernel size from the set {3, 5, 7, 9}. Larger kernel sizes produce stronger blurring effects.

\noindent{\textbf{Compression:}} JPEG compression is applied to the input images by first selecting a random quality factor between 10 and 75. The image is then encoded into JPEG format with this quality factor, introducing lossy compression.

\noindent{\textbf{Cropping:}} A random crop is applied to the input images by selecting a percentage between 5\% and 20\%, determining the crop size in both the x and y directions. The image is then resized to its original dimensions using cubic interpolation.

\section{Implementation Hyperparameter Details}
\label{appendix: hyperparameter}
To ensure the reproducibility of our experimental results, we provide a complete list of all hyperparameters used during training, as shown in Table~\ref{tab: hyperparameter}.

\begin{table}[h]
\centering
\setlength{\tabcolsep}{10.0pt}
\renewcommand{\arraystretch}{1.0}
\begin{tabular}{lcc}
    \toprule
    Hyperparameter & \makecell{Universal\\FakeDetect} & \makecell{GenImage} \\
    \midrule
    seed & 3407 & 3407 \\
    batch\_size  & 4 & 4 \\
    epochs  & 10 & 10 \\
    learning\_rate  & $1 \times 10^{-6}$ & $1 \times 10^{-5}$\\
    lr\_decay\_step  & 1 & 1 \\
    lr\_decay\_factor  & 0.5 & 0.5 \\
    masking\_ratio & (0.0, 0.4) & (0.6, 0.8) \\
    patch\_size & 14 & 14\\
    \bottomrule
\end{tabular}
\caption{Hyperparameters.}
\label{tab: hyperparameter}
\end{table}